\pdfoutput=1

\documentclass[11pt]{article}

\usepackage[final]{coling}

\usepackage{times}
\usepackage{latexsym}

\usepackage[T1]{fontenc}

\usepackage[utf8]{inputenc}

\usepackage{microtype}

\usepackage{hyperref}       
\usepackage{url}            
\usepackage{booktabs}       
\usepackage{amsfonts}       
\usepackage{nicefrac}       
\usepackage{microtype}      
\usepackage{xcolor}         

\usepackage{microtype}
\usepackage{hyperref}
\usepackage{url}
\usepackage{booktabs}
\usepackage{multirow}
\usepackage{enumitem}

\usepackage{comment}
\usepackage{graphicx}
\usepackage{subcaption}

\usepackage{inconsolata}

\usepackage{graphicx}

\usepackage{subcaption}
\usepackage{CJKutf8}

\usepackage{natbib}
\usepackage{booktabs} 
\usepackage{multirow} 

\title{MLaKE: Multilingual Knowledge Editing Benchmark for \\ Large Language Models}

\author{Zihao Wei$^{1,2\ *}$ \quad Jingcheng Deng$^{1,2}$ \thanks{\ \ Equal Contributions}\quad
Liang Pang$^{1}$\thanks{\ \ Corresponding Author} \\
\textbf{Hanxing Ding}$^{1,2}$\quad \textbf{Huawei Shen}$^{1,2}$\quad
\textbf{Xueqi Cheng}$^{1,2}$\\
$^{1}$ Institute of Computing Technology, Chinese Academy of Sciences \\
$^{2}$ University of Chinese Academy of Sciences \\
\texttt{\{weizihao22z, dengjingcheng23s, pangliang\}@ict.ac.cn}\\
\texttt{\{dinghanxing18s, shenhuawei, cxq\}@ict.ac.cn}
}

\begin{document}
\maketitle
\begin{abstract}
The extensive utilization of large language models (LLMs) underscores the crucial necessity for precise and contemporary knowledge embedded within their intrinsic parameters. Existing research on knowledge editing primarily concentrates on monolingual scenarios, neglecting the complexities presented by multilingual contexts and multi-hop reasoning. To address these challenges, our study introduces MLaKE (Multilingual Knowledge Editing), a novel benchmark comprising 4072 multi-hop and 5360 single-hop questions designed to evaluate the adaptability of knowledge editing methods across five languages: English, Chinese, Japanese, French, and German. MLaKE aggregates fact chains from Wikipedia across languages and utilizes LLMs to generate questions and answer. We assessed the effectiveness of current multilingual knowledge editing methods using the MLaKE dataset. Our results show that due to considerable inconsistencies in both multilingual performance and encoding efficiency, these methods struggle to generalize effectively across languages. The accuracy of these methods when editing English is notably higher than for other languages. The experimental results further demonstrate that models encode knowledge and generation capabilities for different languages using distinct parameters, leading to poor cross-lingual transfer performance in current methods. Transfer performance is notably better within the same language family compared to across different families. These findings emphasize the urgent need to improve multilingual knowledge editing methods.\footnote{Our benchmark and source code are available at \url{https://github.com/Hi-archers/MLaKE}.}
\end{abstract}

\section{Introduction}

With the widespread deployment of large language models, ensuring that the knowledge stored in their intrinsic parameters is correct and up-to-date has become a very important topic~\citep{DBLP:conf/iclr/SinitsinPPPB20,DBLP:journals/corr/abs-2311-05656,DBLP:conf/iclr/ChenS24}. knowledge editing serves as a promising solution to this challenge, necessitating timely updates to the knowledge embedded within LLMs~\citep{DBLP:journals/corr/abs-2012-00363,de-cao-etal-2021-editing,DBLP:journals/corr/abs-2402-10612,DBLP:conf/nips/MengBAB22,DBLP:conf/iclr/MitchellLBFM22,DBLP:journals/corr/abs-2402-18150,DBLP:journals/corr/abs-2407-20224}.

Despite considerable efforts devoted to this research field, current studies on knowledge editing typically concentrate on monolingual scenarioi~\citep{DBLP:journals/corr/abs-2306-09212,DBLP:conf/aaai/ZhangZYLSG0H24,DBLP:conf/nips/HuangBZZZSLLZLF23}, where language models are edited and evaluated within the same language~\citep{DBLP:conf/nips/MengBAB22,DBLP:conf/iclr/MengSABB23,DBLP:conf/iclr/MitchellLBFM22}. However, the rapid advancements in large language models (LLMs) have facilitated the widespread adoption of multi-lingual settings~\citep{DBLP:journals/corr/abs-2303-18223,DBLP:journals/corr/abs-2303-04048}. Given this context, the performance of a source-language edited model on other languages remains largely unexplored. 

To address this challenge, our study introduces MLaKE (Multilingual Language Knowledge Editing), a novel benchmark for multilingual multi-hop knowledge editing. MLaKE comprises 5360 single-hop questions and 4072 multi-hop questions designed to test the adaptability of knowledge editing methods across various languages. To ensure the quality and currency of knowledge, we begin by collecting 
fact chains across languages from Wikipedia. Subsequently, we leverage powerful LLM (e.g., ChatGPT) to generate questions in both free-form and multiple-choice formats using fact chains as input. Consequently, The MLaKE dataset comprises single-hop and multi-hop questions in English, Chinese, Japanese, French, and German. This diverse dataset serves as a robust foundation for evaluating the effectiveness of knowledge editing techniques in diverse linguistic environments.

We assessed the effectiveness of various knowledge editing methods on MLaKE, with a focus on their performance in multilingual contexts. The results show that current mainstream knowledge editing methods demonstrate weak generalization in multilingual editing and poor cross-lingual transfer performance. Specifically, these methods not only struggle to edit knowledge in one language while transferring across others, but also exhibits weak performance in the foundational task of editing knowledge within a single language. Moreover, the aforementioned challenges become even more pronounced in multi-hop reasoning scenarios. Our findings underscore the significant impact of language differences on the performance of knowledge editing. To better understand this challenge, we conduct a series of experiments to analyze the effects of linguistic and structural differences on knowledge editing performance.

The main contributions of our work are as:
\begin{itemize}[itemsep=0pt,parsep=0pt]
    \item We collect the multilingual knowledge editing dataset, MLaKE, which comprises 5360 single-hop questions and 4072 multi-hop questions designed to test the adaptability of existing methods across various languages.
    \item We demonstrate that existing knowledge editing methods, when applied to LLMs, suffer from significant shortcomings in multilingual generalization and cross-lingual transfer performance.
    \item Our analysis shows that weak multilingual generalization is primarily due to the models' insufficient multilingual performance and encoding inefficiency. The limited cross-lingual transferability is largely caused by LLMs using distinct parameter sets to encode knowledge for different languages.

    
\end{itemize}

\section{MLaKE: Multi-Lingual Knowledge Editing Benchmark}

\begin{figure*}[htbp]
  \centering
  \includegraphics[width=\textwidth]{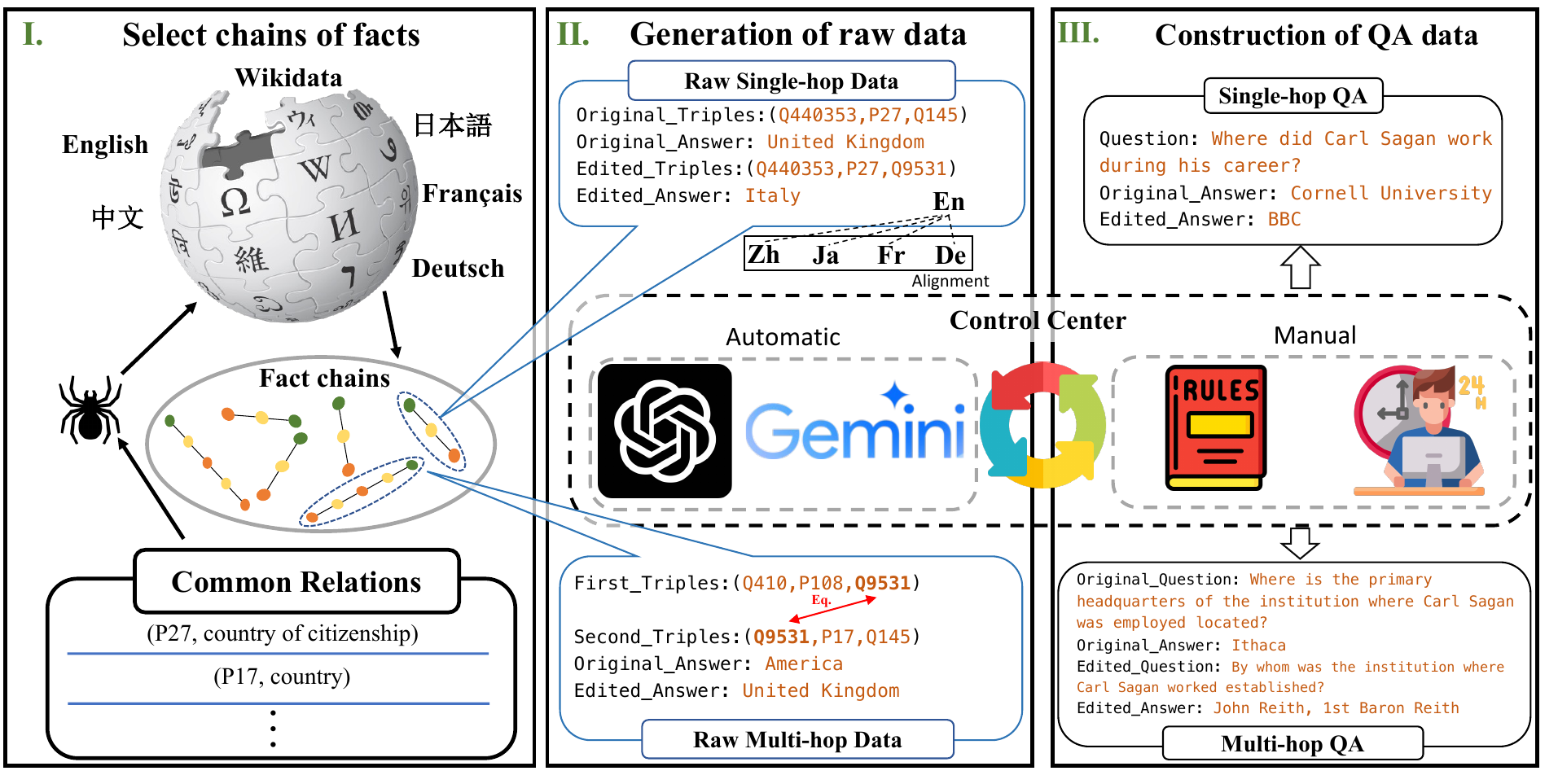}
  \caption{Construction of MLaKE. Firstly, we gather a set of common relations and utilize them to extract single-hop and multi-hop fact chains from Wikidata, encompassing five languages. Then, we combine ChatGPT and manual collaboration to generate edited objects for them, and align the single-hop fact chains. Finally, we utilize the organized raw data to create QA data.}
  \label{fig:construction}
  \vspace{-0.5cm}
\end{figure*}

In our study, we construct the MLaKE (\textbf{M}ulti\textbf{L}ingu\textbf{a}l \textbf{K}nowledge \textbf{E}diting) benchmark, an comprehensive and challenging dataset that encompasses five languages (English, Chinese, Japanese, French, German) and intricate logical structures (single-hop and multi-hop). In this section, we first present the data construction process of MLaKE, followed by a detailed description of the dataset. Lastly, we elaborate on the evaluation settings and metrics utilized.

\subsection{Data Construction of MLaKE}
Figure~\ref{fig:construction} illustrates the construction process of MLaKE, which is primarily composed of three sequential steps: selection and alignment of fact chains, generation of raw data, and construction of question and answer.

\subsubsection{Select chains of facts}
We define fact chains as tuples, where a single-hop fact chain is represented as $(s_1, r_1, o_1)$. In this representation, $s_1$, $r_1$ and $o_1$ denote the subject, relationship, and object of the single-hop, respectively. Similarly, a multi-hop fact chain is expressed as $(s_1, r_1, o_1, r_2, o_2)$, where $r_2$ represents the multi-hop relationship, and $o_2$ signifies the multi-hop object. Notably, the single-hop object is equivalent to the multi-hop subject. 

Inspired by the work of \citet{DBLP:journals/corr/abs-2305-14795}, We gather fact chains by crawling data from Wikipedia\footnote{We collect data from Wikipedia via the Wikidata API: \url{https://www.wikidata.org/w/api.php}}. Initially, we manually create a relational dataset consisting of 43 common relations, the same as the approach taken in previous work \citep{petroni-etal-2019-language, DBLP:conf/nips/MengBAB22}. Subsequently, we collect single-hop fact chains from Wikidata across five languages, leveraging the relational dataset. During this process, we employ rules to ensure that the single-hop fact chain satisfies specific predefined conditions. For instance, to facilitate batch editing, we enforce restrictions that prevent repetitive modifications of the relationship associated with an entity in the single-hop fact chains. For additional filtering rules applied to fact chains, please refer to the Appendix~\ref{app:filter_rules}. To assess the generalizability of the knowledge editing method across languages, we perform alignment on the collected single-hop fact chains. Specifically, we retain only single-hop fact chains that were simultaneously available in all five languages. Given that Wikipedia is written by local communities worldwide, this approach allow us to gather authentic localized expressions. In addition, we continue to collect knowledge from Wikidata for the objects in the single-hop fact chain to form a multi-hop fact chain. Figure~\ref{fig:construction} provides a simple example showcasing this process.

\subsubsection{Generation of raw data}
Once the fact chain is generated, additional data is required to compose the raw data. This additional data primarily encompasses edited answers and answer aliases, as depicted in the Raw Data of Figure~\ref{fig:construction}.

To generate edited answers, we develop instructions to leverage powerful language models like ChatGPT. These instructions ensure the similarity between the edited answer and the original answer while avoiding conflicts with common knowledge within the LLM. For instance, it would be illogical to edit the single-hop knowledge chain ('Carl Sagan', 'employer', 'Cornell University') as ('Carl Sagan', 'employer', 'Glass Cup'). We represent the edited object as $o^*$. The editing process for a single-hop (or multi-hop) fact chain can be expressed as $(s_1, r_1, o_1 \rightarrow o^*_1)$ or $(s_1, r_1, o_1 \rightarrow o^*_1, r_2, o^*_2)$, where $o_2^*$ is the object corresponding to $r_2$ and $o_1^*$.

We initially retrieve the answer alias using the Wikidata API. However, we observe that for certain languages, such as French and German, answers often have diverse variations that are not present in Wikidata. To ensure that these variations do not impact the evaluation (refer to Section~\ref{evaluation}), we design specific instructions for ChatGPT to expand the answer and incorporate appropriate qualifiers into the answer alias.

\subsubsection{Construction of question and answer}
Considering the complexities of inflection in French and German sentence structures, we avoid the template-driven methods often used in prior studies to convert triples into questions. Instead, we utilize ChatGPT to generate fluent and coherent multilingual questions, along with their corresponding answers, based on the collected triplet data in several languages. To minimize potential errors in this transformation process, five experts in the relevant languages were invited to review the model-generated texts, following the criteria detailed in Appendix \ref{app:filter_rules}.

\subsection{Description of the MLaKE}

\paragraph{Dataset statistics} Table~\ref{tab1} summarizes the statistics of the MLaKE dataset. The MLaKE dataset consists of more than 13K samples. We align all single-hop problems and employ them as a means to investigate the generalizability of existing knowledge editing methods across different languages following editing in a single language. Multi-hop problems are not aligned across languages, and we use them to assess the generalizability and transferability of existing editing methods.

\paragraph{Dataset analysis} Figure~\ref{fig1} briefly analyzes the characteristics of the MLaKE dataset. Figure~\ref{fig1}(a) depicts all first relations and their corresponding top 3 second relations, demonstrating the diversity of relations in MLaKE. The majority of questions pertain to nationality, names of individuals and locations, and typically adhere to the following structure: \texttt{"From which nation does Gwendoline Christie hold citizenship?"} (single-hop question) or \texttt{"In which country is the institution where Carl Sagan was employed located?"} (multi-hop question). In Figure~\ref{fig1}(b), we further examine the relation PIDs that account for more than 1\%, mainly including "country of citizenship" (p27), "country" (P17), "continent" (P30), etc. For the corresponding table of relationship PID and relationship label, please refer to the Appendix~\ref{sec:data_details}. Figure~\ref{fig1}(c) depicts the distribution of entities that have an occurrence rate exceeding 0.5\%. The prominent entities in this distribution include 'United Kingdom' (Q145), 'Canada' (Q36), and 'United States of America' (Q30). Figure~\ref{fig1}(d) illustrates the distribution of question lengths. The majority of questions fall within the 10-20 word range, which allows for precise expression of the subject and relationship without the inclusion of extraneous information. To accommodate various answer preferences of Large Language Models (LLMs), we strive to generate multiple aliases for each answer in MLaKE. Figure~\ref{fig1}(e) demonstrates that the majority of answers possess 2-13 aliases, while there are even several answers with more than 20 aliases. The analysis data presented in Figure~\ref{fig1} only includes English samples, both single-hop and multi-hop.

\subsection{Evaluation Metrics}
\label{evaluation}
Diverging from other benchmarks, MLaKE primarily focuses on assessing the generalizability and transferability of knowledge editing methods across multilingual scenarios. For different models, we use corresponding question to guide them to generate answers. We evaluate the accuracy of the question-answering task by determining whether the model-generated sentences contained the correct answer or its aliases.
\section{Experiments}
This section first explains the experimental settings, then analyzes the generalization ability of multilingual knowledge editing and the transfer ability of cross-language knowledge editing, and finally explores the potential reasons that affect the performance of knowledge editing.

\subsection{Experimental Setup}

\paragraph{Language Models}

We use the following two language models in our experiments:  1) \textbf{Vicuna-7B-v1.5} is fine-tuned from Llama-2 using supervised instruction fine-tuning with training data sourced from ShareGPT.com. 2) \textbf{Qwen1.5-7B-Chat} is a transformer-based aligned chat model pre-trained on extensive data, developed by Alibaba Cloud. 

\paragraph{Knowledge Editing Methods}
Building on previous research~\citep{yao-etal-2023-editing,DBLP:journals/corr/abs-2308-07269}, we incorporate four strong knowledge editing methods as baselines: 1) \textbf{MEND}~\citep{DBLP:conf/iclr/MitchellLBFM22} uses small auxiliary networks and gradient decomposition for efficient and localized post-hoc editing. 2) \textbf{ROME}~\citep{DBLP:conf/nips/MengBAB22} employs causal mediation analysis to pinpoint the area for edits, and then adjusts crucial feedforward weights using rank-one model editing. 3) \textbf{MEMIT}~\citep{DBLP:conf/iclr/MengSABB23} extends ROME to edit a large set of facts and facilitate thousands of edits to be executed simultaneously. 4) \textbf{StableKE}~\citep{wei2024stable}, which leverages knowledge augmentation rather than focusing solely on knowledge localization, exhibits stability across various knowledge editing settings.

\paragraph{Evaluation Dimensions}
\begin{itemize}[itemsep=0pt,parsep=0pt]
 \item \textbf{Multilingual generalization} refers to the accuracy of knowledge editing across different languages, especially when edited knowledge is tested using single language.
 \item \textbf{Cross-lingual transferability} refers to the ability of the editing model to apply the knowledge edited in one language to another language, i.e., the accuracy of editing in other languages after performing editing in one language.
\end{itemize}

\begin{table*}[t]
\centering
\resizebox{0.9\linewidth}{!}{
\begin{tabular}{lcccccccccc}
\toprule
\multirow{2}{*}{\textbf{Methods}} & \multicolumn{5}{c}{\textbf{Single-hop QA} (\%) - Vicuna} & \multicolumn{5}{c}{\textbf{Multi-hop QA} (\%) - Vicuna} \\
\cmidrule(r){2-6}\cmidrule(r){7-11} 
& \multicolumn{1}{c}{EN} & \multicolumn{1}{c}{DE} & \multicolumn{1}{c}{FR} & \multicolumn{1}{c}{JA} & \multicolumn{1}{c}{ZH} & \multicolumn{1}{c}{EN} & \multicolumn{1}{c}{DE} & \multicolumn{1}{c}{FR} & \multicolumn{1}{c}{JA} & \multicolumn{1}{c}{ZH} \\
\midrule
MEND    & 0.19 & 0.09 & 0.09 & 0.00 & 0.00 & 0.00 & 0.14 & 0.86 & 0.04 & 0.25 \\
ROME    & 14.93 & 1.59 & 4.38 & 0.09 & 0.28 & 2.44 & 0.00 & 0.57 & 0.11 & 0.25 \\
MEMIT   & 60.73 & 23.79 & 44.22 & 4.94 & 3.73 & 20.70 & 7.11 & 14.12 & 3.19 & 3.30 \\
StableKE & 88.43 & 37.31 & 37.31 & 32.09 & 28.73 & 26.66 & 13.68 & 5.99 & 11.11 & 11.66 \\
\midrule
\multirow{2}{*}{\textbf{Methods}} & \multicolumn{5}{c}{\textbf{Single-hop QA} (\%) - Qwen} & \multicolumn{5}{c}{\textbf{Multi-hop QA} (\%) - Qwen} \\
\cmidrule(r){2-6}\cmidrule(r){7-11} 
& \multicolumn{1}{c}{EN} & \multicolumn{1}{c}{DE} & \multicolumn{1}{c}{FR} & \multicolumn{1}{c}{JA} & \multicolumn{1}{c}{ZH} & \multicolumn{1}{c}{EN} & \multicolumn{1}{c}{DE} & \multicolumn{1}{c}{FR} & \multicolumn{1}{c}{JA} & \multicolumn{1}{c}{ZH} \\
\midrule
ROME    & 44.59 & 21.92 & 20.24 & 0.28 & 33.68 & 15.72 & 4.51 & 5.85 & 0.11 & 16.48 \\
MEMIT   & 75.47 & 59.51 & 51.12 & 20.24 & 43.38 & 41.31 & 19.56 & 18.12 & 9.46 & 27.50 \\
StableKE & 92.82 & 45.34 & 44.96 & 37.31 & 67.44 & 55.37 & 22.02 & 10.56 & 17.05 & 31.94 \\
\bottomrule
\end{tabular}}
\caption{Single-hop and Multi-hop QA performance comparison between Vicuna and Qwen models using various knowledge editing methods across different languages.}
\label{tab:RobustKE}
\vspace{-0.5cm}
\end{table*}

\subsection{Generalization of Multilingual Knowledge Editing}
\label{sec:MultiRobus}

To evaluate the multilingual generalization performance of commonly used knowledge editing methods, we select four widely used knowledge editing methods and evaluated their robustness across five different languages using the Vicuna and Qwen models. Each method are both edited and tested within the same language context. Table~\ref{tab:RobustKE} presents the experimental results, and the main conclusion are as follows.

\textbf{Conclusion 1: All evaluated knowledge editing methods exhibit limited multilingual generalization.} As can be seen in Table~\ref{tab:RobustKE}, all methods demonstrate significantly higher accuracy in English knowledge editing compared to other languages, regardless of the base model used. This may be due to the fact that the quality and scale of English are the highest among the training corpora used by existing models.

\begin{figure}[htbp]
\centering
\includegraphics[width=\linewidth]{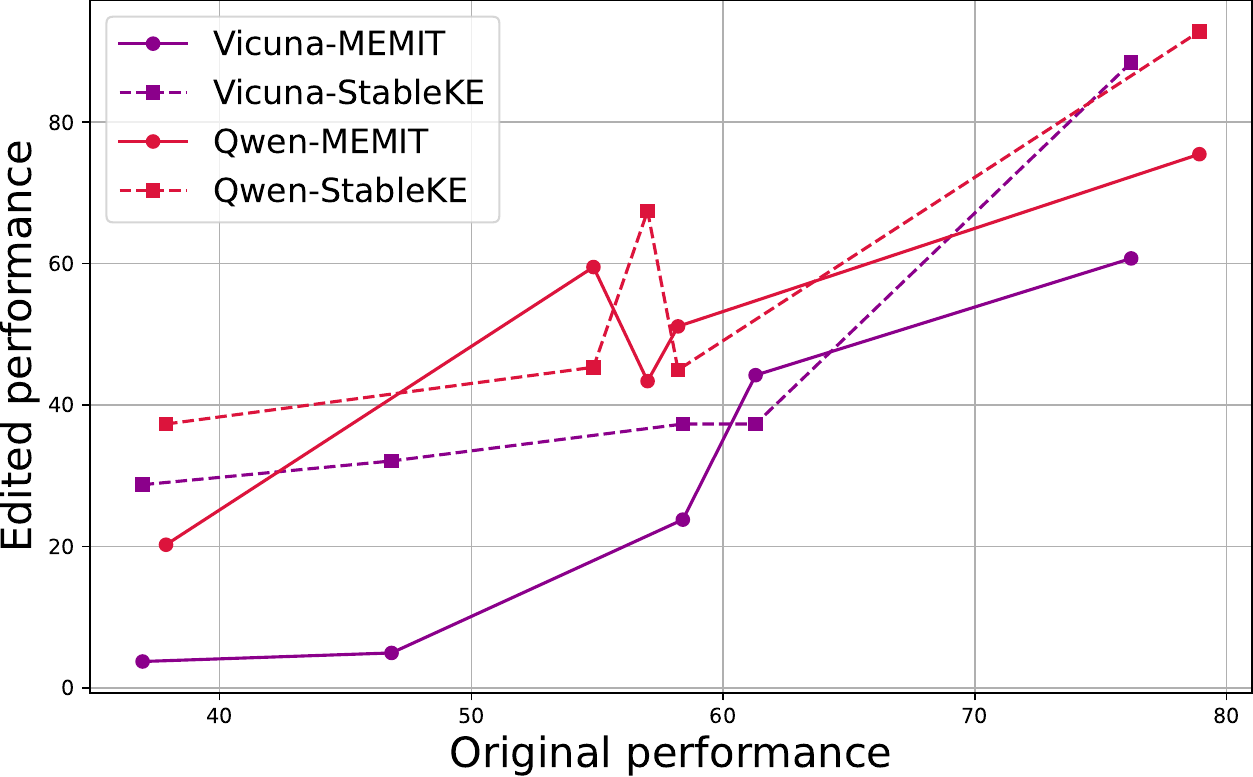}
\caption{The relationship between the original performance and the edited performance of the model. With the same knowledge editing method, the better the model's original performance, the more effective the knowledge editing becomes.}
\label{fig:ori_edit}
\vspace{-0.5cm}
\end{figure}

\textbf{Conclusion 2: The performance of the knowledge editing method is related to the performance of the base model.}
Except for StableKE, the other three methods perform poorly in editing Chinese and Japanese knowledge using the Vicuna model. In contrast, these methods achieve significantly higher accuracy with the Qwen model. The Figure~\ref{fig:ori_edit} shows that the better the base model performs in a particular language, the more effective the knowledge editing method becomes.

Additionally, the table shows that the accuracy of multi-hop knowledge editing is substantially lower than that of single-hop editing. This finding aligns with previous studies focused on English, and our study extends these conclusions to a multilingual scenario.


\begin{figure*}[t]
\centering
\includegraphics[width=\textwidth]{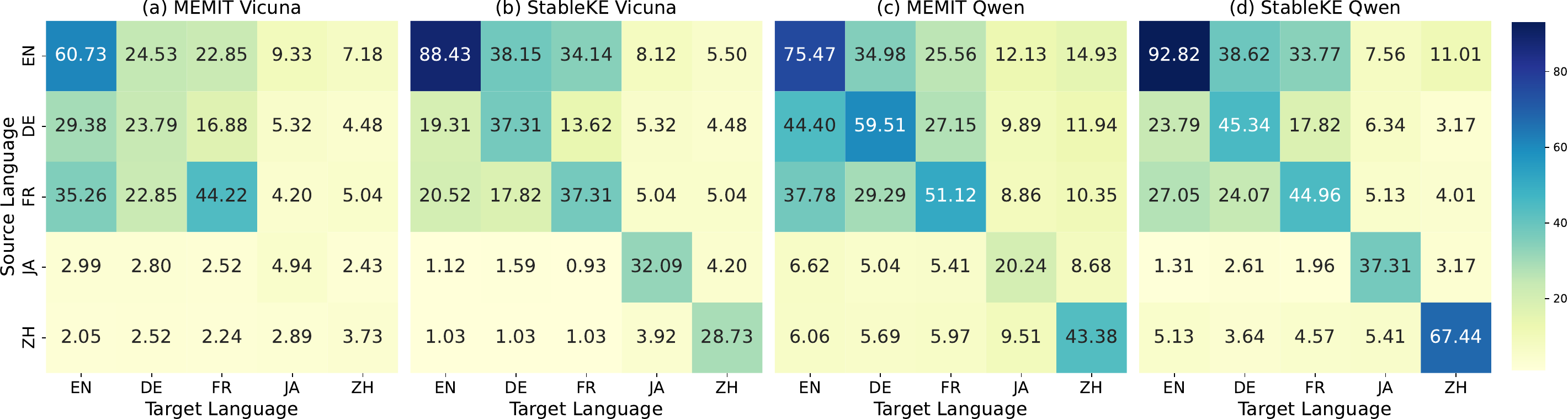}
\caption{Performance of MEMIT and StableKE on different source and edit languages on the vicuna1.5 and Qwen1.5 model.}
\label{fig:hotmap}
\vspace{-0.5cm}
\end{figure*}

\begin{figure}[t]
\centering
\includegraphics[width=\linewidth]{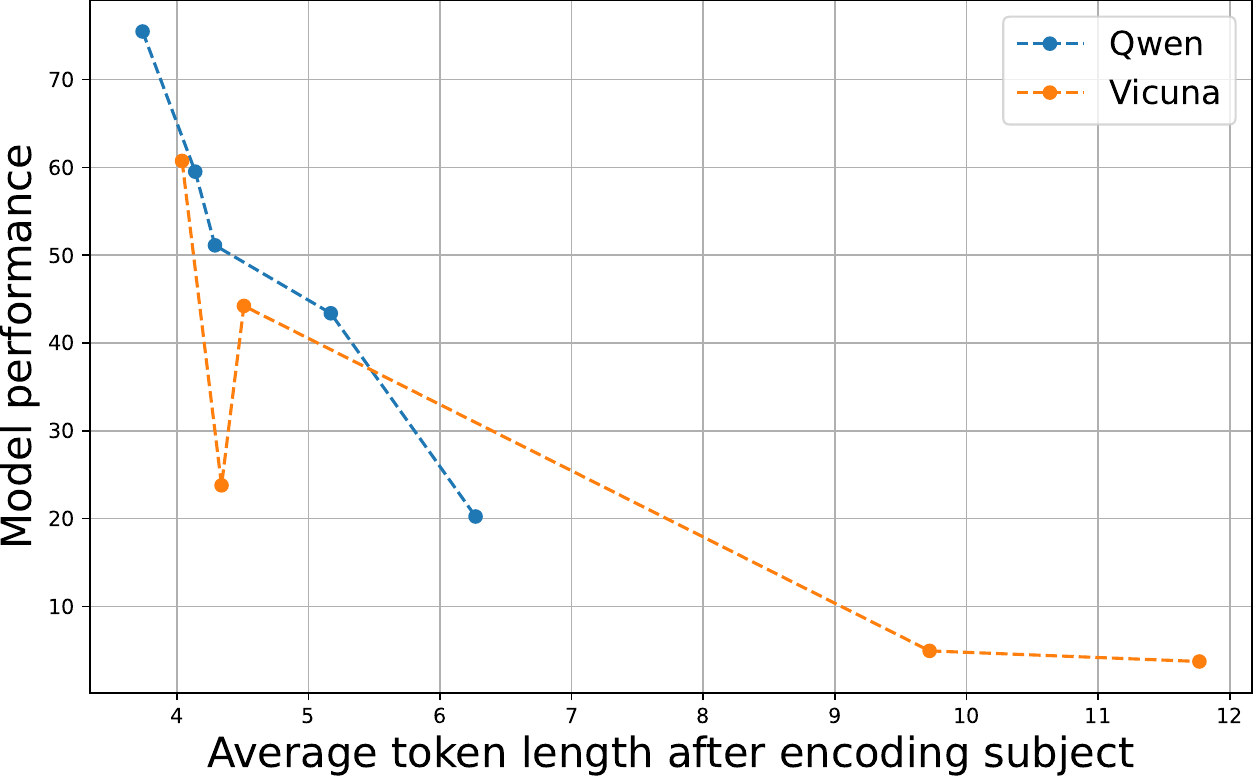}
\caption{Relationship between the average number of tokens needed to encode each subject and the success rate of editing.}
\label{fig:tokenizer}
\vspace{-0.5cm}
\end{figure}

\subsection{Transferability of Cross-Language Knowledge Editing}
\label{sec:Cross_language}

To evaluate the transferability performance of existing knowledge editing methods in cross-language scenarios, we conduct knowledge editing in one language and assessed the editing accuracy in different languages. Our analysis focus on two representative methods: MEMIT and StableKE. Results from additional methods are presented in the appendix, and the main conclusion are as follows.

\textbf{Conclusion 3: All knowledge editing methods have limited cross-language transferability, and the accuracy of the model in answering questions drops significantly when using different languages for reasoning.}
As illustrated in Figures \ref{fig:hotmap}, the editing performance along the diagonal is notably superior, indicating that the cross-language transferability ability of MEMIT and StableKE is limited. Notably, when performing knowledge editing in English on the Vicuna model, the accuracy in other languages was significantly lower than in English. A similar pattern was observed in the Qwen model during both Chinese and English editing. 

\textbf{Conclusion 4: The greater the similarity between the editing language and the reasoning language, the stronger the cross-language transferability.}
According to linguistic classification, English, German, and French all belong to the Indo-European language family~\cite{joseph2005indo}. Although Chinese and Japanese do not belong to the same language family, Japanese has been significantly influenced by Chinese~\cite{handel2008sino,Jinlian2004TheCC}. The figure illustrates a trend: in the top-left corner, cross-linguistic transfer among the three Indo-European languages is notably stronger, whereas the transfer from these languages to Chinese and Japanese is significantly weaker than within the Indo-European family. While there is some linguistic connection between Chinese and Japanese, it is weaker compared to the Indo-European languages, resulting in stronger transfer between Chinese and Japanese than their transfer to other languages, but still weaker than the transfer observed within the Indo-European language family.
\begin{table}[t]
\centering
\begin{tabular}{lcc}
\toprule
\multirow{2}{*}{\textbf{Method}}  & \multicolumn{2}{c}{\textbf{Accuracy}}     \\
\cmidrule(r){2-3}
& \textbf{Use Subject}   & \textbf{No Subject}   \\
\midrule
\textbf{ROME} & 14.93     & 6.16    \\
\textbf{MEMIT}  & 60.73      & 4.91    \\
\bottomrule
\end{tabular}
\caption{Accuracy of MEMIT and ROME with and without using subject.}
\label{table:subject}
\vspace{-0.5cm}
\end{table}

\section{Causes of Multilingual Editing Challenges}

\begin{figure*}[t]
\centering
\includegraphics[width=\textwidth]{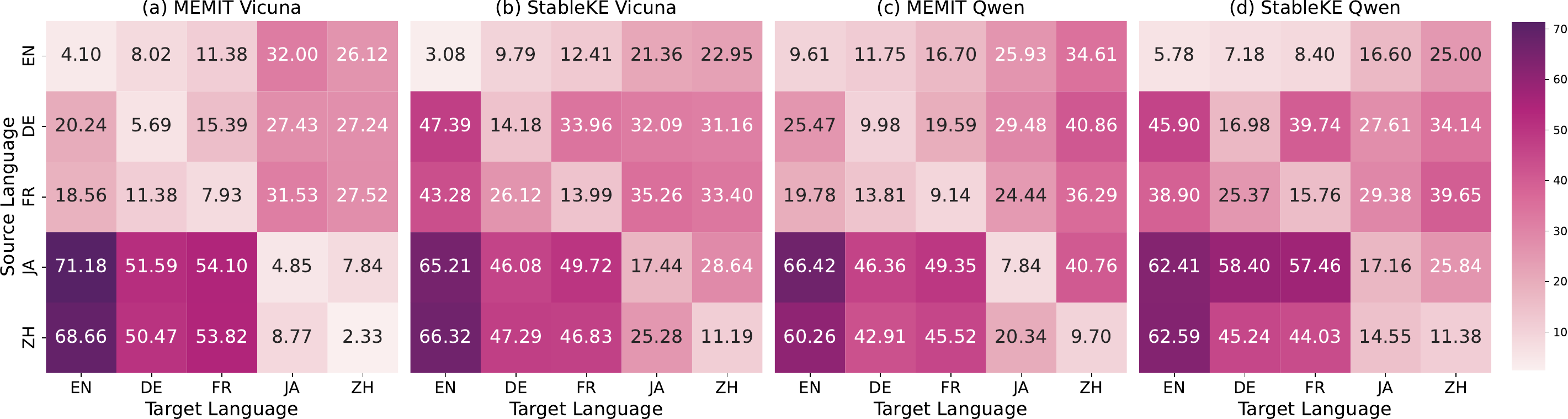}
\caption{Proportion of responses that remain identical to the original, unedited outputs after applying MEMIT and StableKE with different source and edit languages on the Vicuna1.5 and Qwen1.5 model.}
\label{fig:hotmap_raw_vicuna}
\vspace{-0.5cm}
\end{figure*}

We identify two causes contributing to the suboptimal performance of multilingual knowledge editing: significant disparities in the multilingual capabilities of models and the structural independence of multilingual representations. The inconsistency in multilingual performance refers to the model's varying capability and encoding efficiency across different languages, which undermines the generalization of knowledge editing in multilingual contexts. Structural independence in multilingual models suggests that knowledge and abilities in different languages are encoded by distinct parameters. This separation limits the effectiveness of knowledge editing in achieving cross-linguistic transfer.

\subsection{Potential Factors Limiting Multilingual Editing Generalization}
Besides the model's multilingual performance, another crucial factor influencing the generalization ability of multilingual knowledge editing methods is its text encoding performance. To examine the relationship between text encoding performance and the generalization of multilingual knowledge editing, we analyze the ROME and MEMIT methods. Both methods follow the locate-then-edit approach, with their core process focused on identifying the subject’s final token to enhance editing accuracy. However, as shown in Table \ref{table:subject}, when the subject’s final token is replaced with the sentence’s last token, disrupting this process, the knowledge editing accuracy of both MEMIT and ROME declines significantly. Additionally, the Vicuna and Qwen models exhibit substantial differences in encoding efficiency across languages. For instance, as demonstrated in Table \ref{table:tokenizer}, the Vicuna model requires an average of 11.77 tokens to represent a Chinese subject, while only 4.04 tokens are needed for an English subject. As a result, the information density in the final token of a Chinese subject in Vicuna is lower than that of an English subject, negatively impacting target localization accuracy and reducing editing precision. As illustrated in Figure \ref{fig:tokenizer}, as a model’s encoding efficiency decreases (i.e., more tokens are required to encode a target), the success rate of knowledge editing correspondingly diminishes. In addition to the more apparent factor of the model’s multilingual performance influencing generalization, we further demonstrate through experiments that the model’s multilingual text encoding efficiency is another potential factor limiting multilingual editing generalization.

\subsection{Examining Factors Affecting Cross-lingual Knowledge Transfer}
\textbf{Conclusion 5: Knowledge editing has a greater impact on the knowledge of languages that are more closely related to each other.}
To thoroughly investigate the factors influencing cross-lingual transferability in multilingual knowledge editing, we focus on evaluating how edits to knowledge affect content generated in languages that were not directly edited. 
we further analyze how MEMIT and StableKE impacted the models’ ability to generate multilingual text. As shown in Figures \ref{fig:hotmap_raw_vicuna}, our results further illustrate the findings presented in Section \ref{sec:Cross_language}. Knowledge editing exerts a stronger influence on the knowledge of languages closely related to the target language, while exerting a weaker influence on the knowledge of more distantly related languages. For example, when knowledge editing is performed in Chinese or Japanese, the accuracy of the original responses in three Indo-European languages decreases only slightly, indicating minimal disruption. These results suggest that existing knowledge editing methods have a smaller effect on knowledge associated with languages that are less closely related. For instance, when knowledge editing is conducted in Chinese or Japanese, the accuracy of responses in three Indo-European languages declines only marginally, indicating minimal interference. Similarly, when Indo-European languages are used for knowledge editing, the accuracy of original responses in Chinese and Japanese shows a comparable marginal decline, reflecting limited impact. These findings suggest that current knowledge editing methods have a reduced impact on knowledge related to languages that are less closely connected.


\begin{table}[t]
\centering
\begin{tabular}{lccccc}
\toprule
\textbf{Model} & \textbf{EN} & \textbf{DE} & \textbf{FR} & \textbf{JA} & \textbf{ZH} \\
\midrule
\textbf{Vicuna} & 4.04 & 4.34 & 4.51 & 9.72 & 11.77 \\
\textbf{Qwen}   & 3.74 & 4.14 & 4.29 & 6.27 & 5.17 \\
\bottomrule
\end{tabular}
\caption{Average number of tokens needed to encode a subject word by Vicuna and Qwen.}
\label{table:tokenizer}
\vspace{-0.6cm}
\end{table}

\begin{figure*}[t]
\centering
\includegraphics[width=\textwidth]{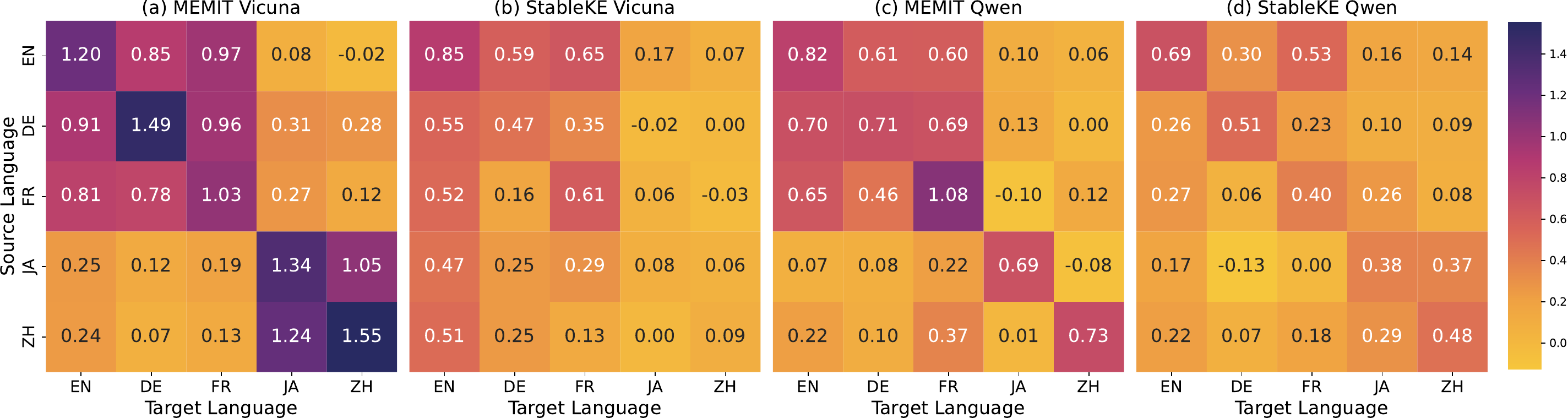}
\caption{Fluency Performance of MEMIT and StableKE on different source and edit languages on the Vicuna1.5 and Qwen1.5 model.}
\label{fig:hotmap_fluency_vicuna}
\vspace{-0.5cm}
\end{figure*}

\begin{figure*}[t]
\centering
\includegraphics[width=\textwidth]{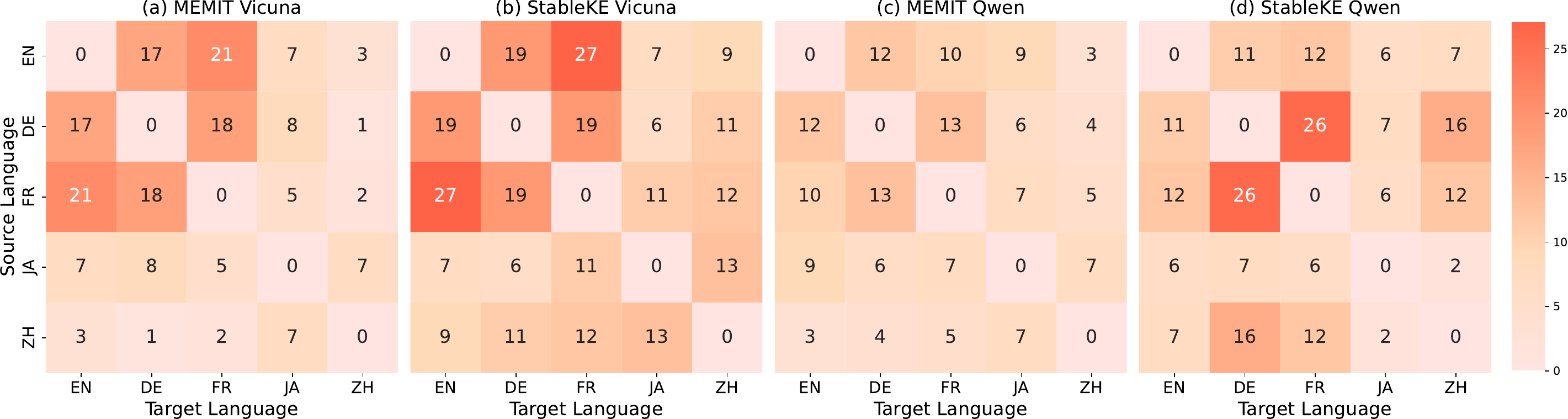}
\caption{Overlap of the Top 100 and 200 Most Impacted Neurons Across Different Languages When Editing Vicuna and Qwen Models Using MEMIT and StableKE.}
\label{fig:memit_mlp}
\vspace{-0.5cm}
\end{figure*}

\textbf{Conclusion 6: Knowledge editing has a greater impact on the generation ability of languages that are more closely related to each other.}
To comprehensively evaluate the impact of multilingual knowledge editing on non-edited languages, we assess its impact on the generative ability of non-edited languages. Without considering the accuracy of the text, we used ChatGPT-4 to evaluate the fluency of the content. As presented in Figure~\ref{fig:hotmap_fluency_vicuna}, knowledge editing minimally affects the generative performance of distantly related languages. Notably, when MEMIT is applied to edit Vicuna's knowledge in Chinese, the model's ability to generate coherent Chinese text deteriorated significantly, resulting in repetitive, incoherent, and meaningless output. In contrast, the impact on English text generation was less pronounced. Specifically, the fluency score for Chinese dropped sharply from 3.84 to 2.29, while the English fluency score declined only slightly from 4.03 to 3.81 following knowledge editing. This trend is observed consistently across multiple language families. We hypothesize that the model encodes knowledge and generative abilities for different languages in distinct parameters, with greater divergence in encoding for more distantly related languages.

To test this hypothesis, we analyze the differences in the multilayer perceptrons (MLPs) that are significantly affected at the same layer during knowledge editing across different languages. Specifically, we identify the neurons that exhibited the largest changes when editing the Vicuna and Qwen models using the MEMIT and StaleKE methods. We then compare the extent of overlap in these neurons across various languages. As illustrated in Figure~\ref{fig:memit_mlp}, the MEMIT method, which focuses on layers 4-8 for editing, operates within a narrower range but exerts a more pronounced effect on the corresponding parameters. We identified the top 100 neurons most impacted by MEMIT and analyzed their cross-linguistic overlap. Additionally, We analyze the overlap between the top 200 neurons primarily influenced by the StaleKE approach, which modifies all parameters of the model. Although StaleKE affects a broader range of parameters, its influence on individual neurons is comparatively smaller. The results indicate that the cross-linguistic overlap of the most impacted neurons is generally low, with all overlaps remaining below 20\%. However, closely related languages demonstrate a higher degree of neuron overlap. For example, languages within the Indo-European family exhibit greater overlap compared to those from different language families. Given that knowledge and capabilities in different languages are encoded in distinct parameters, current knowledge-editing methods face limitations in their ability to generalize edits across languages when applied to a single language.

\section{Conclusion}

In this paper, we explore the effectiveness of current knowledge editing methods in multilingual settings. To this end, we create the MLaKE dataset by extracting knowledge tuples from Wikipedia and generating single-hop and multi-hop questions using ChatGPT. Leveraging MLaKE, we conduct experiments employing various methods and multilingual LLMs to investigate the generalization and transferability of knowledge editing from English and other languages. Our research findings reveal that: (1) Current knowledge-editing methods have limited generalization performance in multilingual settings. Beyond the inconsistent model performance across different languages, experiments suggest that insufficient multilingual encoding may contribute to their weak generalization. (2) The cross-lingual transferability of knowledge-editing methods is similarly constrained. Even within the same language family, transferability remains limited and is especially weak across distinct language families. A key finding from the experiments is that LLMs encode knowledge in different languages using distinct parameter sets, which reduces cross-lingual transfer performance.

\section*{Limitations}

In this paper, we primarily explore knowledge-editing methods, focusing on parameter adjustment. These methods specifically target knowledge encoded in model parameters, and through a comprehensive analysis of their impact on the parameters, we highlight existing limitations in multilingual generalization and cross-lingual knowledge transfer. Our findings provide insights for advancing the development of more robust multilingual knowledge-editing methods. Given time and resource constraints, we do not extensively explore or compare knowledge-editing methods unrelated to parameter changes, like in-context editing.

\section*{Acknowledgements}
This work was supported by the Strategic Priority Research Program of the CAS under Grants No.XDB0680302, the National Natural Science Foundation of China (NSFC) under Grants No. 62276248, and the Youth Innovation Promotion Association CAS under Grants No. 2023111.

\section*{Ethics Statement}
This study thoroughly investigates multilingual knowledge editing in large language models and introduces the Multilingual Knowledge Editing Dataset-MLaKE. Using this dataset, we extensively demonstrate and analyze the performance of existing knowledge editing methods across different languages. The MLaKE dataset reveals notable disparities in current multilingual knowledge editing methods, specifically that the success rate of knowledge editing in English is significantly higher than in other languages. Consequently, MLaKE facilitates the advancement of more equitable and generalizable multilingual knowledge editing methods.

\bibliography{anthology,custom}

\appendix


\section{Evaluating QA Performance of Five Widely Used LLMs on MLaKE}
\label{sec:QA_Perfor}

To evaluate the effectiveness of LLMs in various languages and their content generation capabilities, we assessed five instruction-tuned LLMs: LLaMa-2-7B-chat, Vicuna-7B-v1.5, Qwen1.5-7B-Chat, Gemma-7B-IT, and Mistral-7B-Instruct-v0.2 s~\cite{DBLP:journals/corr/abs-2307-09288,team2024gemma,DBLP:journals/corr/abs-2309-16609,DBLP:journals/corr/abs-2310-06825}. Using the unedited MLaKE data, we assessed two capabilities of these five models. The results for single-hop and multi-hop QA are presented in Table ~\ref{tab:qa_performance}. Our key findings include: 1. The same model can display notable differences in QA performance when tested in different languages. English consistently performs better in both single-hop and multi-hop QA, indicating that LLMs have a deeper grasp of English. 2. Qwen1.5-7B-Chat demonstrates a more consistent performance across various languages than other LLMs, achieving especially strong results in Chinese and Japanese. We attribute this advantage to its pre-training corpus, which includes a diverse range of languages~\cite{DBLP:journals/corr/abs-2309-16609}. 3. Conversely, Vicuna-7B-v1.5 shows markedly improved generative capabilities in Chinese and Japanese. This enhancement is mainly attributed to the inclusion of Chinese and Japanese data in the ShareGPT instruction-tuning dataset, which strengthens its ability to generate responses in these languages. 4. Gemma-7B-IT shows significant differences in accuracy in free-form QA compared to other models. Notably, it falls behind LLaMa-2-7B-Chat by 27.8\% in single-hop Free form QA and 15.11\% in multi-hop Free form QA. This gap is mainly attributed to the RLHF optimization in the Gemma model, which often leads to its refusal to generate responses.

\begin{CJK*}{UTF8}{gbsn}

\begin{table*}[t]
\centering
\resizebox{\textwidth}{!}{
\begin{tabular}{lcccccccccc}
\toprule
\multirow{2}{*}{\textbf{Models}} & \multicolumn{5}{c}{\textbf{Single-hop QA} (\%)} & \multicolumn{5}{c}{\textbf{Multi-hop QA} (\%)} \\
\cmidrule(r){2-6}\cmidrule(r){7-11}
& \multicolumn{1}{c}{EN} & \multicolumn{1}{c}{DE} & \multicolumn{1}{c}{FR} & \multicolumn{1}{c}{JA} & \multicolumn{1}{c}{ZH} & \multicolumn{1}{c}{EN} & \multicolumn{1}{c}{DE} & \multicolumn{1}{c}{FR} & \multicolumn{1}{c}{JA} & \multicolumn{1}{c}{ZH} \\
\midrule
LLaMa-2-7B-Chat    &   78.17     &   59.05  &  63.62  &  0.0  &   0.0   &  49.94  &   25.99    &   29.92   &    0.0   &  0.0 \\
Vicuna-7B-v1.5  &   76.21   &   58.40   &  61.29  &   46.83    &  36.94  &  47.60    &   28.50   &   26.98  &  26.50   &  21.32 \\
Qwen1.5-7B-Chat    & 78.92  &  54.85   &   58.21  &  37.87   &   57.00    &  60.81  &  30.46    &   30.56   &   30.51   &  53.55  \\
Gemma-7B-IT   &   50.37   &  42.07  & 42.35  &  43.00   &   34.33   &  34.83   &  12.94    &  22.51   &   24.26    &  22.11 \\
Mistral-7B-Ins.-v0.2    &  83.21   &   60.73   &   65.95  &  26.77   &  28.17   &  62.12   &    40.39   &  37.21   &   18.85   &  29.21  \\
\bottomrule
\end{tabular}}
\caption{Single-hop and Multi-hop Free-form QA performance of five LLMs on MLaKE.}
\label{tab:qa_performance}
\end{table*}

\section{Implementation Details}
All experiments are conducted on a single NVIDIA A800 GPU (80GB). We re-implement MEND, ROME and MEMIT using EasyEdit\footnote{\url{https://github.com/zjunlp/EasyEdit}}~\citep{DBLP:journals/corr/abs-2308-07269} with default settings. Additionally, we reproduce the results of StableKE by utilizing their official repository\footnote{\url{https://github.com/Hi-archers/StableKE}}. In the case of MEND, MEMIT, and StableKE, a batch size of 100 was employed for actual editing. As for ROME, which does not support batch editing, a batch size of 1 was used, and 100 samples were iteratively edited before testing.

\section{Data Details}
\label{sec:data_details}
In this section, we present the rules employed for filtering the data, followed by the instructions utilized for constructing the data.

\begin{table*}[htbp]
\centering
\begin{tabular}{lccccc}
\toprule
\multirow{2}{*}{\textbf{Statistics}}  & \multicolumn{5}{c}{\textbf{Language}}     \\
\cmidrule(r){2-6}
& EN   & ZH   & JA   &   FR   &  DE   \\
\midrule
\textbf{\# single-hop questions} & 1072     & 1072    & 1072     & 1072     & 1072     \\
\textbf{\# multi-hop questions} (\texttt{original})  & 916      & 760     & 849    &  782    & 765  \\
\textbf{\# multi-hop questions} (\texttt{edited})  &   1024   &  798  &   909   &  701    & 731  \\
\midrule
\textbf{\# single-hop entities}  &   602   &  596  &   596   &   596   & 594  \\
\textbf{\# multi-hop entities} (\texttt{original})  & 1004  &  858  &  968    &   988   &   994 \\
\textbf{\# multi-hop entities} (\texttt{edited})  &   1303   &  1053  &   1253  &   1267   & 1285  \\
\midrule
\textbf{\# relations}  &   43   &  43  &   43  &   43   &  43 \\
\bottomrule
\end{tabular}
\caption{Data statistics of MLaKE. Multi-hop questions are not aligned cross five languages, so we mark them with \texttt{original} and \texttt{edited} respectively. EN denotes English, ZH denotes Chinese, JA denotes Japanese, FR denotes French, and DE denotes German.}
\label{tab1}
\vspace{-0.6cm}
\end{table*}

\begin{figure*}[t]
\begin{minipage}[b]{.52\textwidth}
    \centering
    \subfloat[][First relations (inner circle) and their top 3 second relations.]{\label{fig:sunburst_chart}\includegraphics[width=1\linewidth]{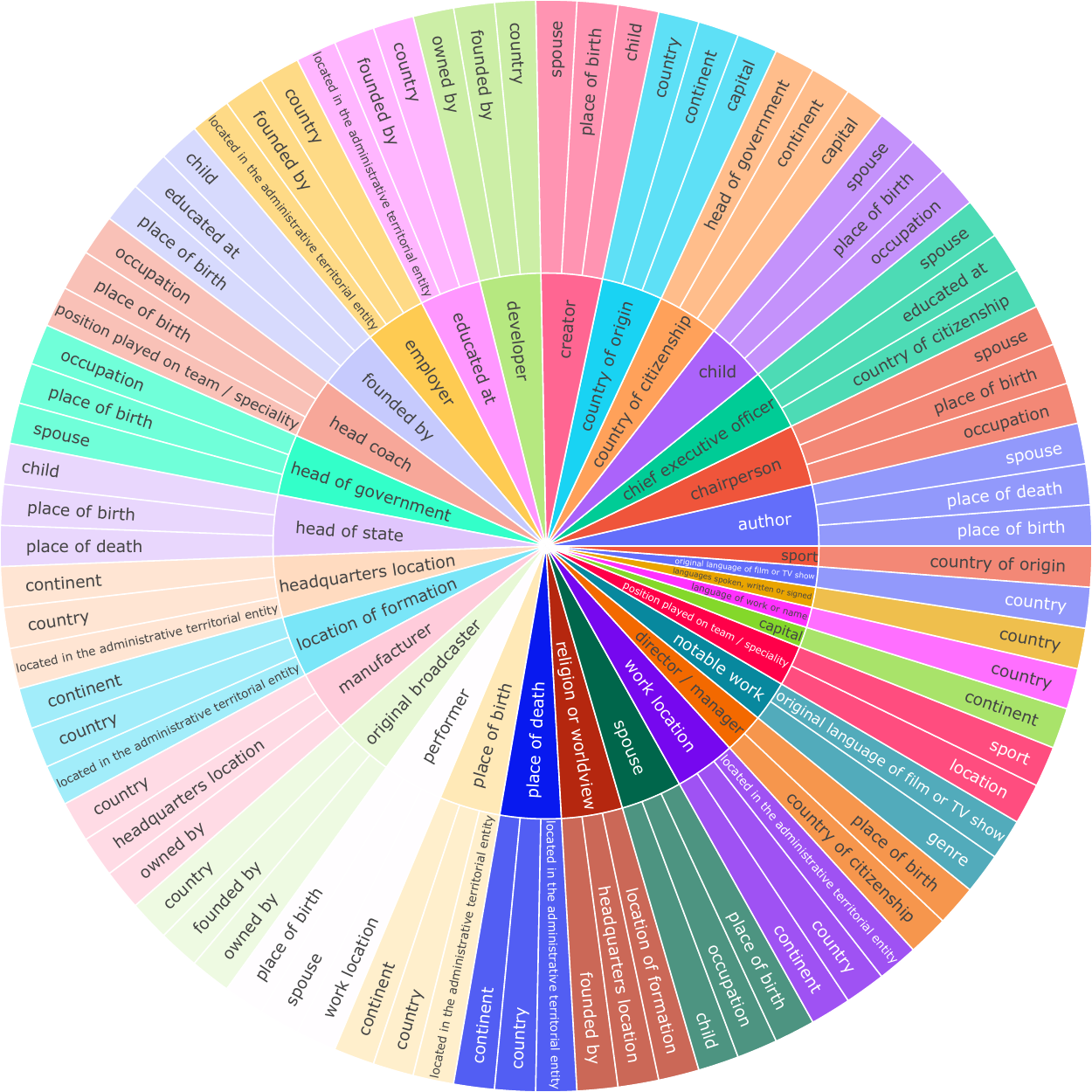}}
\end{minipage}
\hfill
\begin{minipage}[b]{.45\textwidth}
    \begin{flushright}
    \subfloat[][Frequency distribution  of relations (only occurrences $>$1\% are shown).]{\label{fig:relation_bar}\includegraphics[width=0.9\linewidth]{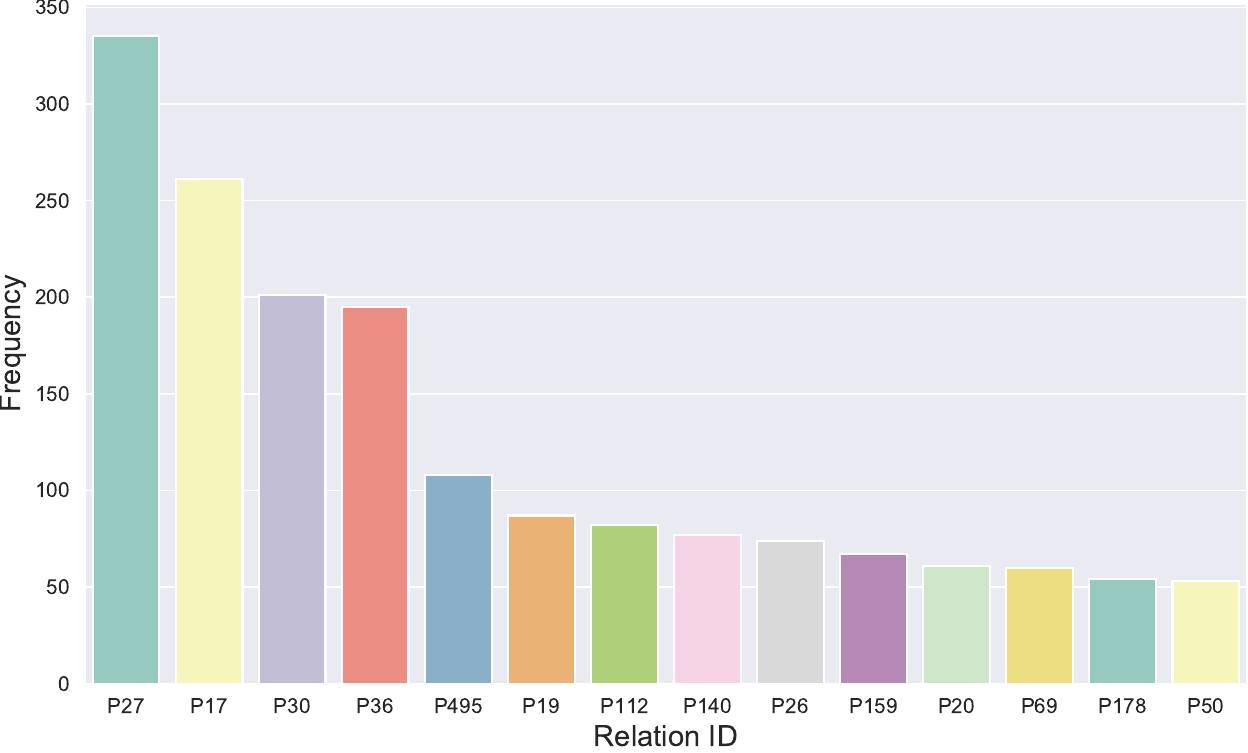}}
    
    \subfloat[][Frequency distribution of entities (only occurrences $>$0.5\% are shown).]{\label{fig:entity_distribution}\includegraphics[width=0.9\linewidth]{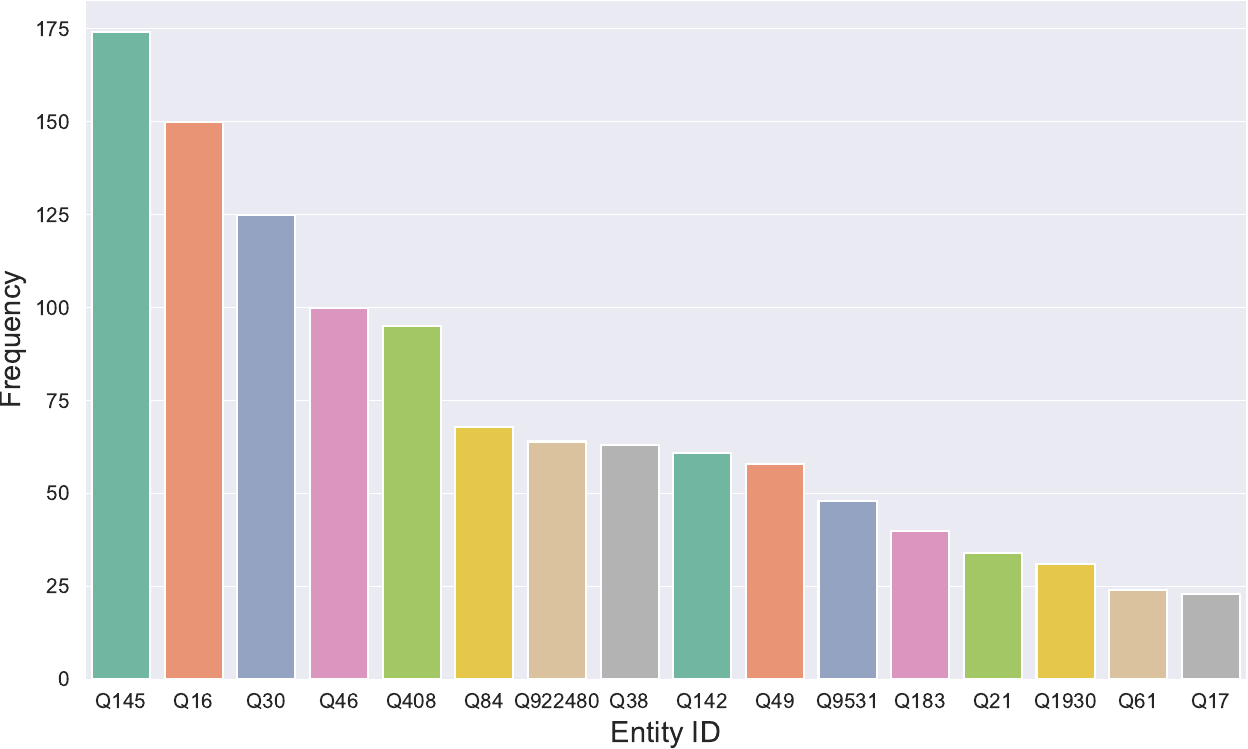}}
    \end{flushright}
\end{minipage}

\begin{minipage}[b]{\textwidth}
    \centering
    \subfloat[Distribution of the length of questions.]{\label{fig:question_length}\includegraphics[width=0.48\linewidth]{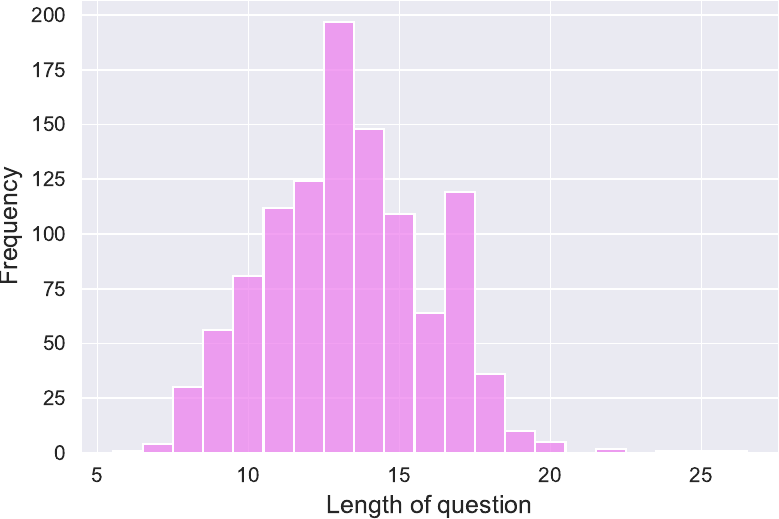}}
    \hfill
    \subfloat[Distribution of the number of edited answer aliases.]{\label{fig:aliases_distribution}\includegraphics[width=0.48\linewidth]{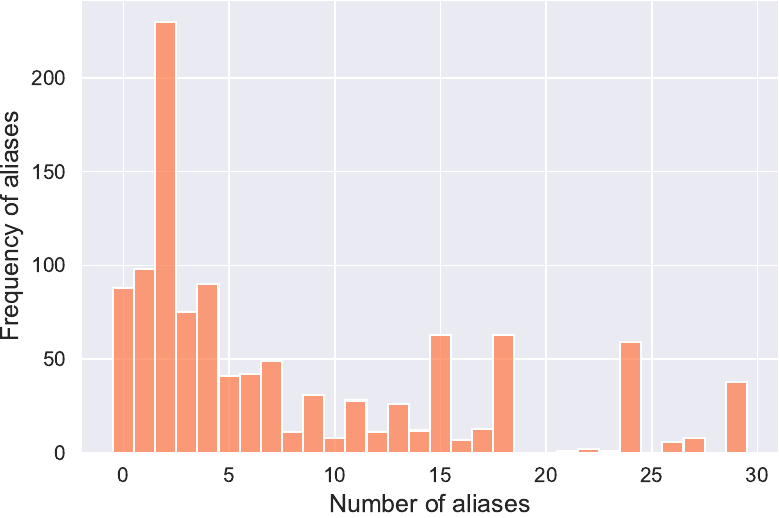}}
\end{minipage}
\caption{Analysis of MLaKE Dataset. \textbf{(a)} We illustrate the connections between the first relations (inner circle) and their corresponding second relations (outer circle). \textbf{(b)} We depict the distribution of relations occurring more than 1\%. \textbf{(c)} We visualize the distribution of entities occurring more than 0.5\%. \textbf{(d)} We present the distribution of question lengths. \textbf{(e)} We display the distribution of the number of edited answer aliases.}
\label{fig1}
\vspace{-0.5cm}
\end{figure*}

\subsection{Filter rules}
\label{app:filter_rules}
We hired five professionals, each specializing in one of the five languages used in this paper, to conduct a comprehensive data cleaning exercise on the dataset according to the following guidelines:

\begin{enumerate}[itemsep=2pt,topsep=0pt,parsep=0pt]
    \item All single-hop data has corresponding five language representations, which is the alignment operation mentioned in the main text. It serves as the foundational basis for conducting experiments on the generalization of multilingual knowledge editing.
    \item Among all data, the relationship corresponding to an entity cannot be modified repeatedly. To facilitate evaluation and avoid knowledge conflicts, we ensure that the fact chain of an entity is not modified multiple times. This is particularly relevant considering that several knowledge editing methods support batch editing capabilities.
    \item The dataset is free of toxic information, including content related to politics, violence, or pornography.
    \item Fact chains, particularly multi-hop ones, are free from circular dependencies.
    \item Identify and eliminate any semantic or syntactic errors in the questions and answers generated by the model.
\end{enumerate}

\subsection{Instructions}
Tables ~\ref{tab:single_hop insructions} and ~\ref{tab:multi_hop insructions} respectively display the prompts that guided ChatGPT in generating single-hop and multi-hop questions, as well as their corresponding answers.

\begin{table*}
\centering
\resizebox{0.95\linewidth}{!}{
\begin{tabular}{lp{396pt}}
\toprule
\multicolumn{2}{c}{\textbf{English instructions for single-hop data}}\\
\midrule
\multirow{4}{*}{Question Generation Instruction} & Given the Wikidata knowledge triplet structure [subject, relation, object] where subject is $s_1$, relation is $r_1$, use this information to guide ChatGPT in creating a question that aims to identify the $o_1$ of the triplet. Your challenge is to create a detailed question that prompts LLM to identify the $r_1$ of $s_1$, without giving away the $o_1$ or making any reference to it.\\
\\ 
\multirow{4}{*}{Answer Generation Instruction} & Using the provided Wikidata knowledge triple [$s_1$, $r_1$, $o_1$], craft a concise answer to the question (question). Your response should clearly link the $s_1$ with the $o_1$ as the answer, without delving into additional details or context. The aim is to directly address the query with the information given in the triple.\\
\midrule
\multicolumn{2}{c}{\textbf{Chinese instructions for single-hop data}}\\
\midrule
\multirow{4}{*}{Question Generation Instruction} &  给定Wikidata知识三元组[subject, relation, object]，subject是$s_1$，relation是$r_1$,。根据这些信息，设计一个中文问题，旨在根据$s_1$和$r_1$询问三元组中的object，即$o_1$。问题应详细且具体，让便让LLM准确的回答出$s_1$的$r_1$，同时避免直接提到或暗示$o_1$。\\\\
\multirow{3}{*}{Answer Generation Instruction} &根据给定的Wikidata知识三元组[$s_1$, $r_1$, $o_1$], 就问题(question)形成一个简洁的回答。您的回答应明确地将$s_1$与$o_1$联系起来作为答案，避免深入其他细节或背景。请力求简明扼要，利用特定的三元组关系，确保回答简洁且直接相关。请直接生成结果，不要说无关的内容。\\
\midrule
\multicolumn{2}{c}{\textbf{Japanese instructions for single-hop data}}\\
\midrule
\multirow{5}{*}{Question Generation Instruction} &  Wikidataの知識トリプレット構造[subject, relation, object]が与えられた場合、ここでsubjectは$s_1$、relationは$r_1$です。この情報を使用して、トリプレットの$o_1$を特定することを目的とした質問をChatGPTが作成するよう導いてください。あなたの挑戦は、$o_1$ を明かしたり、それに言及したりすることなく、$s_1$の$r_1$を特定するようLLMに促す詳細な質問を作成することです。\\\\
\multirow{3}{*}{Answer Generation Instruction} & ウィキデータの知識トリプル [$s_1$, $r_1$, $o_1$] を考慮すると、 、質問「(question)」に対する正確な応答を作成します。あなたの答えは、「$r_1$」によって提供されるコンテキストを通じて、「$s_1$」を「$r_1$」に直接関連付ける必要があります。トリプルで示される特定の関係を活用した簡単な説明を目指し、応答が簡潔で質問に直接関連していることを確認します。 \\
\midrule
\multicolumn{2}{c}{\textbf{French instructions for single-hop data}}\\
\midrule
\multirow{3}{*}{Question Generation Instruction} &  Étant donné la structure de triplet de connaissances Wikidata [subject, relation, object] où subject est $s_1$, relation est $r_1$, utilisez cette information pour créer une question avec ChatGPT pour identifier l'élément $o_1$ du triplet. Votre défi est de créer une question détaillée qui incite LLM à identifier le $r_1$ de $o_1$, sans révéler le $o_1$ ou faire référence à celui-ci.\\\\
\multirow{5}{*}{Answer Generation Instruction} &  Étant donné le triplet de connaissances Wikidata [$s_1$, $r_1$, $o_1$], formulez une réponse concise à la question (question). Votre réponse devrait clairement lier le $s_1$ avec le $o_1$ comme réponse, sans entrer dans des détails ou contextes supplémentaires. Visez une explication simple qui tire parti de la relation spécifique dénotée par le triplet, en vous assurant que la réponse est succincte et directement pertinente à la question. Veuillez générer les résultats directement et ne dites pas de contenu hors sujet.\\
\midrule
\multicolumn{2}{c}{\textbf{German instructions for single-hop data}}\\
\midrule
\multirow{4}{*}{Question Generation Instruction} &  Angesichts der Struktur eines Wissens-Tripels in Wikidata (subject, relation, object), wobei 'subject' $s_1$ ist, 'relation' $r_1$ ist, nutzen Sie diese Informationen, um ChatGPT bei der Erstellung einer Frage zu leiten, die darauf abzielt, das $o_1$ des Tripels zu identifizieren. Ihre Herausforderung besteht darin, eine detaillierte Frage zu formulieren, die LLM dazu anregt, das $r_1$ von $s_1$ zu identifizieren, ohne das $o_1$ preiszugeben oder darauf Bezug zu nehmen.\\\\
\multirow{3}{*}{Answer Generation Instruction} &  Verwendung des bereitgestellten Wikidata-Wissenstripels ($s_1$, $r_1$, $o_1$), verfassen Sie eine prägnante Antwort auf die Frage (question). Ihre Antwort sollte $s_1$ eindeutig mit $o_1$ als Antwort verknüpfen, ohne auf zusätzliche Details oder den Kontext einzugehen. Ziel ist es, mit den im Tripel enthaltenen Informationen direkt auf die Anfrage einzugehen.\\
\bottomrule
\end{tabular}}
\caption{Instructions required to generate single-hop data in five languages.}
\label{tab:single_hop insructions}
\end{table*}

\begin{table*}
\centering
\resizebox{0.95\linewidth}{!}{
\begin{tabular}{lp{396pt}}
\toprule
\multicolumn{2}{c}{\textbf{English instructions for multi-hop data}}\\
\midrule
\multirow{4}{*}{Question Generation Instruction} &  Given the Wikidata triples: ($s_1$, $r_1$, $o_1$) and ($o_1$, $r_2$, x2), craft a multi-hop question in natural English about $s_1$ that explicitly involves the relationships $r_1$ and $r_2$. The question must ensure there is no implicit or explicit reference to or information leakage about $s_1$, leading to an inquiry about x2 without revealing $s_1$.\\\\
\multirow{5}{*}{Option Generation Instruction} & Given the information: '{question}' and the known facts: ($s_1$, $r_1$, $o_1$) and ($o_1$, $r_2$, x2), please generate a correct option A and provide three incorrect but plausible options B, C, and D. Ensure that all options are presented in a sentence, not just single words or phrases. The incorrect options should be related enough to the correct answer to pose a challenge, but there's no need to mention the intermediary connecting entity ($s_1$) or any other detailed information.\\
\midrule
\multicolumn{2}{c}{\textbf{Chinese instructions for multi-hop data}}\\
\midrule
\multirow{2}{*}{Question Generation Instruction} &     请根据以下Wikidata知识三元组：（$s_1$, $r_1$, $o_1$）和（$o_1$, $r_2$, x2），用流畅的中文提出一个关于$s_1$的多跳问题，用于通过$s_1$查询得到x2，该问题必须明确涉及关系$r_1$和$r_2$，从而引出关于x2的询问。(问题要确保没有对$s_1$的直接或间接引用或信息泄露)。\\\\
\multirow{3}{*}{Option Generation Instruction} & 给定信息：'{question}'，已知事实包括：($s_1$, $r_1$, $o_1$)和($o_1$, $r_2$, x2)。请根据这些信息，生成一个正确的选项A，并提供三个错误但听起来合理的选项B、C和D。请确保这些选项是以完整的句子形式提出的，而不仅仅是单个词或短语，错误选项应该与正确答案有一定的关联度，使问题具有一定的挑战性，但无需提及中间的连接实体($s_1$)或其他详细信息。\\
\midrule
\multicolumn{2}{c}{\textbf{Japanese instructions for multi-hop data}}\\
\midrule
\multirow{2}{*}{Question Generation Instruction} &  次のWikidataの知識トリプルに基づいて、$s_1$についての1つの質問を作成してください：（$s_1$, $r_1$, $o_1$）および（$o_1$, $r_2$, x2）。この質問は、$r_1$と$r_2$の関係を通じて$s_1$からx2への論理的な連鎖を明確にし、$s_1$に関する直接または間接的な参照を避けなければなりません。\\\\
\multirow{4}{*}{Option Generation Instruction} &    情報「{question}」と既知の事実「$s_1$, $r_1$, $o_1$」および「$o_1$, $r_2$, x2」に基づき、正しい選択肢Aを生成し、不正解ながらも妥当な選択肢B、C、Dを3つ提供してください。すべての選択肢は、単語やフレーズだけでなく、文章で提示する必要があります。不正解の選択肢は、正解と十分関連していて挑戦となる必要がありますが、中間の接続エンティティ「$s_1$」や他の詳細な情報に言及する必要はありません。\\
\midrule
\multicolumn{2}{c}{\textbf{French instructions for multi-hop data}}\\
\midrule
\multirow{4}{*}{Question Generation Instruction} &  Veuillez poser une question multi-sauts en français fluide basée sur les triplets de connaissances Wikidata suivants : ($s_1$, $r_1$, $o_1$) et ($o_1$, $r_2$, x2). La question doit concerner $s_1$ et servir à interroger pour obtenir x2, en mentionnant explicitement les relations $r_1$ et $r_2$ pour introduire une question sur x2. (La question doit s'assurer qu'il n'y ait aucune référence directe ou indirecte à $s_1$ ou de fuite d'informations à son sujet).\\\\
\multirow{6}{*}{Option Generation Instruction} &  Informations fournies: '{question}' Les faits connus incluent: ($s_1$, $r_1$, $o_1$) et ($o_1$, $r_2$, x2). Sur la base de ces informations, veuillez générer une option correcte A et trois options incorrectes mais raisonnables B, C et D. Assurez-vous que les options sont présentées sous forme de phrases complètes et pas seulement de mots ou d'expressions simples. Les mauvaises options doivent avoir un certain degré de pertinence par rapport à la bonne réponse pour rendre la question difficile, mais sans mentionner l'entité de connexion entre les deux ($s_1$) ou d'autres détails.\\
\midrule
\multicolumn{2}{c}{\textbf{German instructions for multi-hop data}}\\
\midrule
\multirow{5}{*}{Question Generation Instruction} &  Ich möchte eine deutsche Multi-Hop-Frage formulieren, basierend auf den gegebenen Wikidata-Wissens-Tripeln: ($s_1$, $r_1$, $o_1$) und ($o_1$, $r_2$, x2). Die Frage soll flüssig in deutscher Sprache gestellt werden und es ermöglichen, durch Abfrage von {x0} die Information über x2 zu erhalten. Dabei muss die Frage die Beziehungen {r1} und {r2} explizit einbeziehen, ohne jedoch direkte oder indirekte Hinweise auf die Brückenentität {x1} zu geben.\\\\
\multirow{5}{*}{Option Generation Instruction} &  Gegebene Informationen: '{question}' Zu den bekannten Fakten gehören: ($s_1$, $r_1$, $o_1$) und ($o_1$, $r_2$, x2). Generieren Sie auf der Grundlage dieser Informationen bitte eine korrekte Option A und drei falsche, aber vernünftig klingende Optionen B, C und D. Stellen Sie sicher, dass die Optionen als vollständige Sätze und nicht nur als einzelne Wörter oder Phrasen dargestellt werden. Die falschen Optionen sollten einen gewissen Grad an Relevanz für die richtige Antwort haben, um die Frage herausfordernd zu gestalten, ohne jedoch die verbindende Entität dazwischen zu erwähnen ($s_1$) oder andere Details.\\
\bottomrule
\end{tabular}}
\caption{Instructions required to generate multi-hop data in five languages.}
\label{tab:multi_hop insructions}
\end{table*}

\section{Related Work}
We introduce recent datasets and knowledge editing methods in this section.

\subsection{Knowledge Editing Datasets}
\label{Knowledge Editing Dataset}
Current research in knowledge editing datasets predominantly focuses on monolingual contexts. For instance, RIPPLEEDITS~\citep{DBLP:journals/corr/abs-2307-12976}, with its 5,000 instances of factual edits in English, serves as a pivotal benchmark designed to examine the ripple effects in knowledge editing processes. Similarly, MQuAKE delves into English multi-hop queries~\citep{DBLP:journals/corr/abs-2305-14795}, evaluating how edits influence intricate chains of knowledge. KEbench offers a thorough assessment of the stability of various knowledge editing methods using a tree-structured dataset in English. In contrast, the works of Bi-ZsRE~\citep{DBLP:journals/corr/abs-2309-08952} and MzsRE~\citep{DBLP:journals/corr/abs-2312-13040} extend to a multilingual knowledge editing dataset by translating the English Zero-Shot Relation Extraction (ZsRE)~\citep{DBLP:conf/conll/LevySCZ17} dataset into various languages. Nonetheless, such endeavors in translation might introduce discrepancies in entity alignment, thereby possibly diminishing the quality of datasets ~\citep{DBLP:journals/corr/abs-2309-08952}.

\subsection{Knowledge Editing Methods}

Current knowledge editing methods can be classified into four paradigms based on knowledge storage and learning techniques: locate-then-edit, memory-based models, meta-learning, and knowledge augmentation. \textbf{Locate-Then-Edit} involves identifying and updating a subset of model parameters associated with the edited knowledge. For instance, \citet{dai-etal-2022-knowledge} manipulates `knowledge neurons' (KN) within pretrained transformers to update facts. Similarly, \citet{DBLP:conf/nips/MengBAB22} introduces a method called Rank-One Model Editing (ROME) that modifies key feedforward weights to edit factual associations in LLMs.  However, both ROME and KN are limited to modifying one piece of knowledge at a time. To address this limitation, \citet{DBLP:conf/iclr/MengSABB23} extended the capabilities of ROME and developed MEMIT, enabling batch modification of knowledge simultaneously~\citep{DBLP:journals/corr/abs-2301-04213,DBLP:journals/corr/abs-2311-09053}. \textbf{Memory-based Model} facilitates editing by introducing a small auxiliary model or additional parameters within the MLP layer, without altering the parameters of the original model. SERAC is a method that modifies knowledge by optimizing a counterfactual model~\citep{DBLP:conf/icml/MitchellLBMF22}. On the other hand, T-Patcher achieves knowledge editing by integrating a few trainable neuron patches into the MLP layer~\cite{DBLP:conf/iclr/HuangSZZR023}. In addition, CALINET leverages the characteristics of MLP layers to directly calibrate factual knowledge within LLMs~\citep{dong-etal-2022-calibrating}. \textbf{Meta-learning} utilizes a hypernetwork specifically designed to handle the necessary alterations for manipulating knowledge within the MLP layers of models. KnowledgeEditor~\citep{de-cao-etal-2021-editing} make use of hypernetworks to facilitate efficient edits in language models. MEND~\citep{DBLP:conf/iclr/MitchellLBFM22} introduces auxiliary networks and enables scalable edits by decomposing gradients. \textbf{Knowledge augmentation} mainly includes StableKE ~\citep{wei2024stable} method enhances the stability and effectiveness of knowledge editing in large language models through two automated knowledge augmentation strategies: Semantic Paraphrase Enhancement and Contextual Description Enrichment.

\newpage
\section{Multilingual Knowledge Editing Experiment}
We present Figure~\ref{fig:hotmap} in table form, as shown in Table~\ref{tab:MEMIT_Hotmap} and Table~\ref{tab:StableKE_Hotmap}. The accuracy of English responses after editing in Chinese and Japanese exhibits a notable decrease compared to editing in French and German. This highlights the substantial influence of language disparities on the performance of cross-language knowledge editing.

\begin{table*}[ht]
\centering
\resizebox{\textwidth}{!}{
\begin{tabular}{lcccccccccc}
\toprule
\multirow{2}{*}{\textbf{Source Language}} & \multicolumn{5}{c}{\textbf{Single-hop Free-form QA} (\%)}    & \multicolumn{5}{c}{\textbf{Multi-hop Free-form QA} (\%)}  \\
\cmidrule(r){2-6}\cmidrule(r){7-11} 
& \multicolumn{1}{c}{EN} & \multicolumn{1}{c}{DE} & \multicolumn{1}{c}{FR} & \multicolumn{1}{c}{JA} & \multicolumn{1}{c}{ZH} & \multicolumn{1}{c}{EN} & \multicolumn{1}{c}{DE} & \multicolumn{1}{c}{FR} & \multicolumn{1}{c}{JA} & \multicolumn{1}{c}{ZH} \\
\midrule
EN    &\textbf{60.73} &24.53 &22.85 &9.33 &7.18 &\textbf{20.70} &15.18 &14.12 &10.01 &9.13  \\
DE   &\textbf{29.38} &23.79 &16.88 &5.32 &4.48 &\textbf{17.97} &7.11 &10.13 &10.45 &7.86  \\
FR    &35.26 &22.85 &\textbf{44.22} &4.20 &5.04 &\textbf{21.29} &9.03 &14.12 &9.90 &8.49  \\
JA    &2.99 &2.80 &2.52 &\textbf{4.94} &2.43 &\textbf{9.67} &6.84 &8.42 &3.19 &3.17  \\
ZH    &2.05 &2.52 &2.24 &2.89 &\textbf{3.73} &\textbf{10.94} &6.16 &7.28 &3.74 &3.30  \\
\bottomrule
\end{tabular}}
\caption{The capability of MEMIT to edit knowledge in the source language and to generate accurate responses in a different target language.}
\label{tab:MEMIT_Hotmap}
\end{table*}

\begin{table*}[ht]
\centering
\resizebox{\textwidth}{!}{
\begin{tabular}{lcccccccccc}
\toprule
\multirow{2}{*}{\textbf{Source Language}} & \multicolumn{5}{c}{\textbf{Single-hop Free-form QA} (\%)}    & \multicolumn{5}{c}{\textbf{Multi-hop Free-form QA} (\%)}  \\
\cmidrule(r){2-6}\cmidrule(r){7-11} 
& \multicolumn{1}{c}{EN} & \multicolumn{1}{c}{DE} & \multicolumn{1}{c}{FR} & \multicolumn{1}{c}{JA} & \multicolumn{1}{c}{ZH} & \multicolumn{1}{c}{EN} & \multicolumn{1}{c}{DE} & \multicolumn{1}{c}{FR} & \multicolumn{1}{c}{JA} & \multicolumn{1}{c}{ZH} \\
\midrule
EN    &\textbf{88.43} &38.15 &34.14 &8.12 &5.50 &\textbf{26.66} &17.10 &15.41 &6.21 &8.69  \\
DE    &19.31 &\textbf{37.31} &13.62 &5.32 &4.48 &\textbf{17.77} &13.68 &12.98 &6.97 &8.14 \\
FR    &20.52 &17.82 &\textbf{37.31} &5.04 &5.04 &15.14 &9.03 &5.99 &7.60 &\textbf{15.51}  \\
JA    &1.12 &1.59 &0.93 &\textbf{32.09} &4.20 &9.67 &6.84 &7.28 &6.34 &\textbf{11.11}  \\
ZH    &1.03 &1.03 &1.03 &3.92 &\textbf{28.73} &8.59 &5.61 &6.13 &\textbf{11.66} &7.59  \\
\bottomrule
\end{tabular}}
\caption{The capability of StableKE to edit knowledge in the source language and to generate accurate responses in a different target language.}
\label{tab:StableKE_Hotmap}
\end{table*}


\end{CJK*}

\begin{table*}[ht]
    \centering
    \small
    \begin{tabular}{cl}
\toprule
\multicolumn{2}{c}{\textbf{English Single Hop Free Form QA}}\\
\midrule
$\mathcal{C}$ & Question : Who is the mastermind behind the character Hannibal Lecter?\\
 & original answer : Thomas Harris\\
 & original answer aliases: William Thomas Harris III\\
 & edit answer : Spede Pasanen\\
 & edit answer aliases: None\\
\midrule

\multicolumn{2}{c}{\textbf{German Single Hop Free Form QA}}\\
\midrule
$\mathcal{C}$ & Question : Wer ist der Schöpfer des Charakters, der als Hannibal Lecter bekannt ist?\\
 & original answer : Thomas Harris\\
 & original answer aliases: None\\
 & edit answer : Spede Pasanen\\
 & edit answer aliases: None\\

\midrule

\multicolumn{2}{c}{\textbf{French Single Hop Free Form QA}}\\
\midrule
$\mathcal{C}$ & Question : Qui est l'esprit derrière le personnage de Hannibal Lecter ?\\
 & original answer : Thomas Harris\\
 & original answer aliases: None\\
 & edit answer : Spede Pasanen\\
 & edit answer aliases: None\\

\midrule

\multicolumn{2}{c}{\textbf{Japanese Single Hop Free Form QA}}\\
\midrule
$\mathcal{C}$ & \begin{CJK*}{UTF8}{min}Question : ハンニバル・レクターの物語を生み出した作家は誰ですか？\end{CJK*}\\
 & \begin{CJK*}{UTF8}{min}original answer : トマス・ハリス\end{CJK*}\\
 & \begin{CJK*}{UTF8}{min}original answer aliases: トーマス・ハリス\end{CJK*}\\
 & \begin{CJK*}{UTF8}{min}edit answer : スペデ・パサネン\end{CJK*}\\
 & \begin{CJK*}{UTF8}{min}edit answer aliases: None\end{CJK*}\\

\midrule

\multicolumn{2}{c}{\textbf{Chinese Single Hop Free Form QA}}\\
\midrule
$\mathcal{C}$ & \begin{CJK*}{UTF8}{gbsn}Question : 汉尼拔·莱克特的作品是由哪位作者创作的？\end{CJK*}\\
 & \begin{CJK*}{UTF8}{gbsn}original answer : 托马斯·哈里斯\end{CJK*}\\
 & \begin{CJK*}{UTF8}{gbsn}original answer aliases: None\end{CJK*}\\
 & \begin{CJK*}{UTF8}{gbsn}edit answer : 斯佩德·帕萨宁\end{CJK*}\\
 & \begin{CJK*}{UTF8}{gbsn}edit answer aliases: None\end{CJK*}\\

\bottomrule
\end{tabular}

    \caption{Qualitative examples of the generated multi-hop questions on.}
    \label{tab:q_examples}
\end{table*}





\end{document}